\documentclass[9pt,shortpaper,twoside,web]{ieeecolor}
\usepackage{generic}
\usepackage{cite}
\usepackage{amsmath,amssymb,amsfonts}
\usepackage{algorithmic}
\usepackage{graphicx}
\usepackage{textcomp}
\usepackage{multirow}
\usepackage[T1]{fontenc}
\usepackage{xtab,longtable}
\usepackage{tabularray}

\def\BibTeX{{\rm B\kern-.05em{\sc i\kern-.025em b}\kern-.08em
    T\kern-.1667em\lower.7ex\hbox{E}\kern-.125emX}}
\begin{document}
\title{Multitask Deep Learning for Accurate Risk Stratification and Prediction of Next Steps for Coronary CT Angiography Patients}
\author{Juan Lu, Mohammed Bennamoun \IEEEmembership{Senior Member, IEEE}, Jonathon Stewart, Jason K. Eshraghian \IEEEmembership{Member, IEEE}, Yanbin Liu,  Benjamin Chow, Frank M. Sanfilippo and Girish Dwivedi
\thanks{Manuscript submitted on 21 April. This project was supported by the Western Australian Health Translation Network's Health Service Translational Research project grant, and the funding source had no involvement in the study.}
\thanks{J. L. is with 
the Medical school and Harry Perkins Institute of Medical Research and School of Computer Science and Software Engineering, the University of Western Australia, Perth, WA 
6009 Australia, (e-mail: juan.lu@uwa.edu.au).}
\thanks{M. B. is with the School of Computer Science and Software Engineering, the University of Western Australia, Perth, WA 
6009 Australia (e-mail: mohammed.bennamoun@uwa.edu.au).}
\thanks{J. S. is with 
the Medical School and Harry Perkins Institute of Medical Research, the University of Western Australia, Perth, WA, 6009 (e-mail: jonathon.stewart@uwa.edu.au).}
\thanks{J. E. was with School of Computer Science and Software Engineering, the University of Western Australia Perth, WA, 6009 Australia. He is now with the Department of Electrical and Computer Engineering, University of California, Santa Cruz, CA, 95064 USA (e-mail: jeshragh@ucsc.edu).}
\thanks{Y. L.
is with the School of Computing, Australian National University, Canberra, ACT 2601 Australia (e-mail: csyanbin@gmail.com).}
\thanks{B. C. is with the Division of Cardiology and Departments of Medicine, University of Ottawa Heart Institute, Ottawa, ON K1Y 4W7 (e-mail: bchow@ottawaheart.ca).}
\thanks{F. S. is with the School of Population and Global Health, the University of Western Australia, Perth, WA 6009 Australia (e-mail: frank.sanfilippo@uwa.edu.au).}
\thanks{G. D. is with 
the Medical School and Harry Perkins Institute of Medical Research, the University of Western Australia. He is also with the Department of Cardiology, Fiona Stanley Hospital, Perth, WA, 6154 (e-mail: girish.dwivedi@perkins.uwa.edu.au).}
}
\maketitle

\begin{abstract}
Diagnostic investigation has an important role in risk stratification and clinical decision making of patients with suspected and documented Coronary Artery Disease (CAD). However, the majority of existing tools are primarily focused on the selection of gatekeeper tests, whereas only a handful of systems contain information regarding the downstream testing or treatment. We propose a multi-task deep learning model to support risk stratification and down-stream test selection for patients undergoing Coronary Computed Tomography Angiography (CCTA). The analysis included 14,021 patients who underwent CCTA between 2006 and 2017. Our novel multitask deep learning framework extends the state-of-the art Perceiver model to deal with real-world CCTA report data. 
Our model achieved an Area Under the receiver operating characteristic Curve (AUC) of 0.76 in CAD risk stratification, and 0.72 AUC in predicting downstream tests. 

Our proposed deep learning model can accurately estimate the likelihood of CAD and provide recommended downstream tests based on prior CCTA data. In clinical practice, the utilization of such an approach could bring a paradigm shift in risk stratification and downstream management. Despite significant progress using deep learning models for tabular data, they do not outperform gradient boosting decision trees, and further research is required in this area. However, neural networks appear to benefit more readily from multi-task learning than tree-based models. This could offset the shortcomings of using single task learning approach when working with tabular data.
\end{abstract}


\section{Introduction}
\label{sec:introduction}
Coronary Artery Disease (CAD) is one of the leading causes of death around the globe~\cite{ralapanawa_epidemiology_2021}. 
It accounts for millions of death annually and is a huge global economic burden~\cite{james_global_2018:articleetal}.  Therefore, providing precise diagnostic tests and prevention support can have a substantial influence on reducing CAD-related mortality and save billions for the economy~\cite{savira_impact_2021}.

Coronary Computed Tomography Angiography (CCTA) has evolved as an accurate method for the non-invasive evaluation of CAD, and to direct subsequent treatment~\cite{alaref_machine_2020}. 
It provides anatomical information on the severity of stenosis. 
Invasive Coronary Angiography (ICA) and revascularization may be carried out after the CCTA for patients with obstructive coronary artery stenosis~\cite{task_force_members_2013_2013}. 
However, the recent PROMISE trail\footnote{The PROMISE (Prospective Multicenter Imaging Study for Evaluation of Chest Pain) is a large-scale study that was conducted to evaluate the use of non-invasive tests, such as CCTA and stress tests, in the diagnosis and management of patients with stable chest pain.
The study was designed to compare the diagnostic accuracy and clinical outcomes of these non-invasive tests with those of the traditional invasive test, Invasive Coronary Angiography (ICA)} revealed that CCTA-guided treatment resulted in more ICAs and revascularizations without improving clinical outcomes compared to functional testing~\cite{douglas_outcomes_2015}. 
Functional tests such as exercise stress electrocardiography (EST), single-photon emission computerized tomography (SPECT), and positron emission tomography (PET) evaluate the degree of blood flow restriction (or ischemia) to the heart muscle due to blockages in the coronary arteries.
It has been shown that the integration of functional tests with CCTA test will identify patients who would benefit most from further invasive investigation and revascularization~\cite{van_rosendael_impact_2017}.

\begin{figure}[b]%
\centerline{\includegraphics[width=\columnwidth]{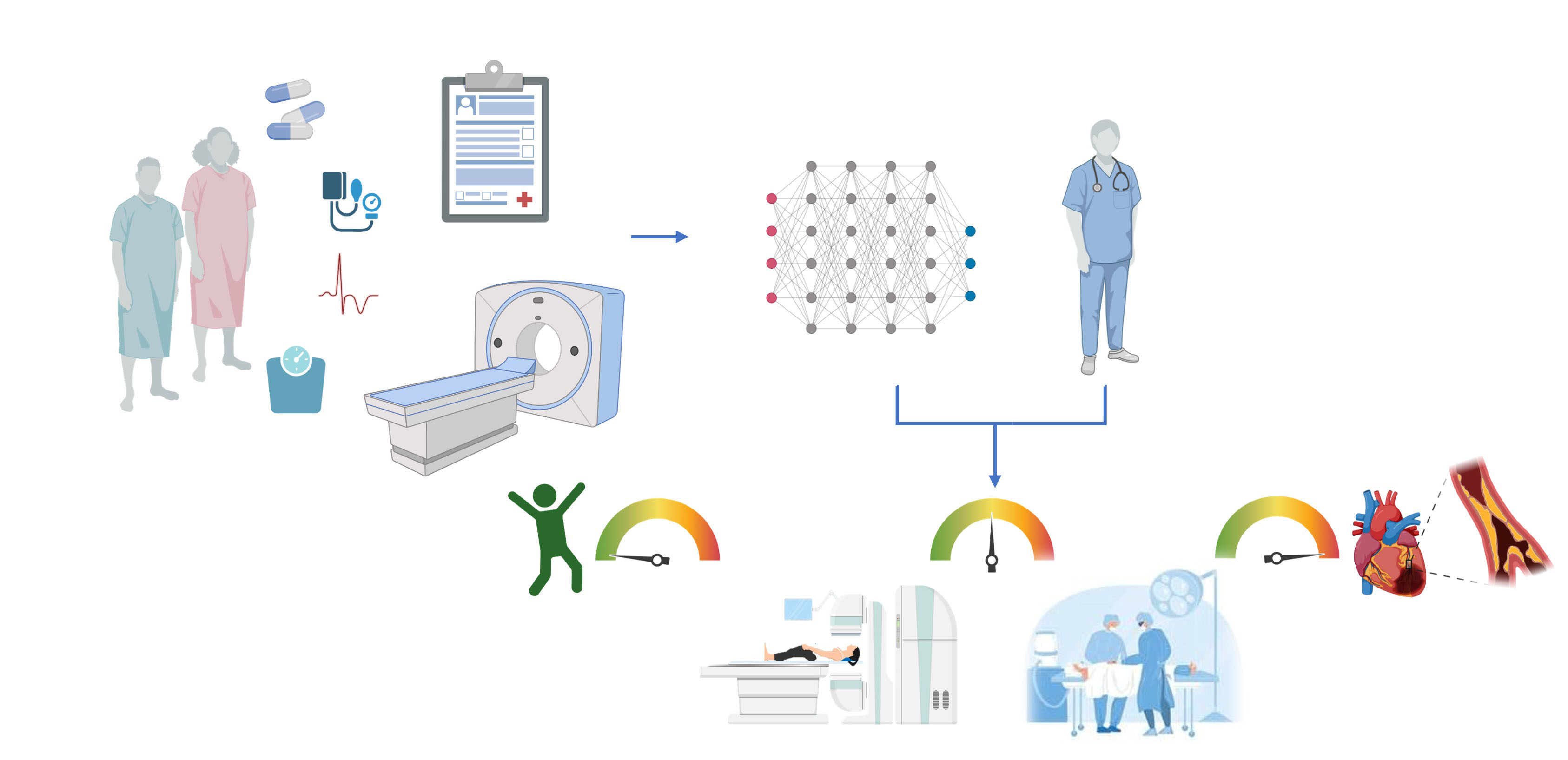}}
\caption{Study Framework. A multitask deep learning neural network was implemented with patients' information (demographics, vital sign, medications, comorbidity history, and coronary computed tomography angiography test readings) jointly predict risk of coronary artery disease and down-streaming investigations or treatments. }\label{Framework}
\end{figure}

It is common in clinical practice to use a validated pretest probability model for predicting the probability of CAD to inform the choice of diagnostic test. 
Most existing tools primarily focus on identifying the gatekeeper test (i.e., a diagnostic test or screening tool that is used to determine whether a patient should be referred for further testing or treatment), while only a few provide information on follow up testing or treatment options~\cite{morgan-hughes_downstream_2021, fyyaz_computed_2020}. 
Meanwhile, it has been observed that the overuse of diagnostic imaging is on the rise in clinical practices causing a significant stress on health budget and fundings~\cite{huang_overuse_2015}. 
As a result, current guidelines emphasize the importance of using risk stratification tools and pre-test probability assessment tools (i.e., to assess the likelihood of disease) prior to initiation of downstream testing. However, their lack of general applicability has been identified as a limitation. Multiple studies have shown that these risk assessment models perform sub-optimally in certain groups of patients~\cite{fihn_2014_2014,task_force_members_2013_2013}. Patients have varying underlying conditions that require personalized approach. Furthermore, differences in derivation (variations in the way these models are developed, such as the use of different imaging modalities, cut-off values for defining CAD risk, and data processing techniques) often occur and may further limit the effectiveness of these tools. 
Machine learning models are ideally suited for addressing these challenges. It could suggest an optimized probability based on the most recent locally relevant data as input. 
Therefore, our aim was to develop a machine learning model, using the CCTA diagnostic test reports of patients as input. These data are collected as tabular data.

As a subfield of machine learning, deep neural networks has achieved dazzling success in a wide variety of areas, such as computer vision and natural language processing using transformer-based architecture~\cite{dosovitskiy_image_2021, brown_language_2020}. 
For tabular data, tree-based machine learning or ensemble models are reported as the best algorithms~\cite{shwartz-ziv_tabular_2021, schmidhuber_deep_2015, borisov_deep_2022},
as these models have been found to be effective in delivering good performance with less training time and are easily explainable. 
However, tree-based models can be sensitive to small changes in the input and have limitations when dealing with limited labeled data when advanced semi-supervised learning techniques are used and in handling high dimensional data~\cite{tanha_semi-supervised_2017, devlin_bert_2019}. 
Conversely, deep learning models have their own strengths, such as the ability to handle both single-task and multi-task learning of image or text inputs, or multi-modal input, including tabular data~\cite{xu_multimodal_2022}. In this paper, we sought to apply deep learning to construct a multi-task learning model on tabular data.

It is logical to assume that the downstream diagnosis of each patient is primarily based on their CAD risk, so it makes a very strong case to consider the CAD risk and downstream diagnosis together.
In this scenario, multi-task learning~\cite{caruana_multitask_1997} is a viable solution as it allows to exploit useful information from these two related learning tasks.

Multitask learning, which is a promising field in deep learning, aims to improve the accuracy of learning each task by leveraging the useful information shared across multiple tasks.
It has been empirically and theoretically demonstrated that learning multiple tasks can result in better performance than learning them individually~\cite{zhang_overview_2018}. 
In this research, we aimed to introduce a multi-task deep learning model for identifying patients who may benefit from invasive coronary investigation and revascularization. Our multi-task deep learning model was designed to learn from patient demographics, clinical history, and CCTA test readings, in order to predict patients' risk of CAD and their down-streaming investigations and treatment options. 

The contributions of this paper are summarized as following: 
\begin{itemize}

\item The introduction of TabPerceiver, a simple and less computationally expensive transformer-based deep learning model that can handle multi-modal data. 

\item  The multi-task TabPerceiver model automatically associates downstream diagnosis and treatment with patient CAD risk.

\item  Demonstrating that TabPerceiver achieved better performance than gradient boosting decision trees but there is no universally superior solution between GBDT and deep learning models for tabular data. 

\end{itemize}
 
\section{Methods}
\label{sec:methods}
\subsection{Study Data}

\begin{figure}[t]%
\centerline{\includegraphics[width=\columnwidth]{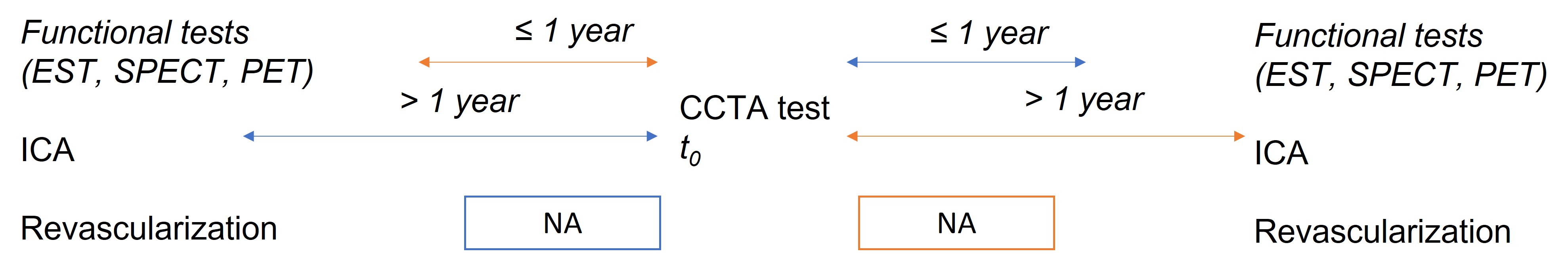}}
\caption{Timeline of cohort selection and output definition. We identified patients who had CCTA, and $t_0$ is the date of their index CCTA. Patients who have undergone previous tests or treatments within 1 year before $t_0$ are excluded. The eligible down-streaming tests should be conducted within 1 year after $t_0$. Patients with no follow-up records are excluded since their outcomes are unknown. }\label{Timeline}
\end{figure}

\begin{figure}[b]%
\centering
\centerline{\includegraphics[width=\columnwidth]{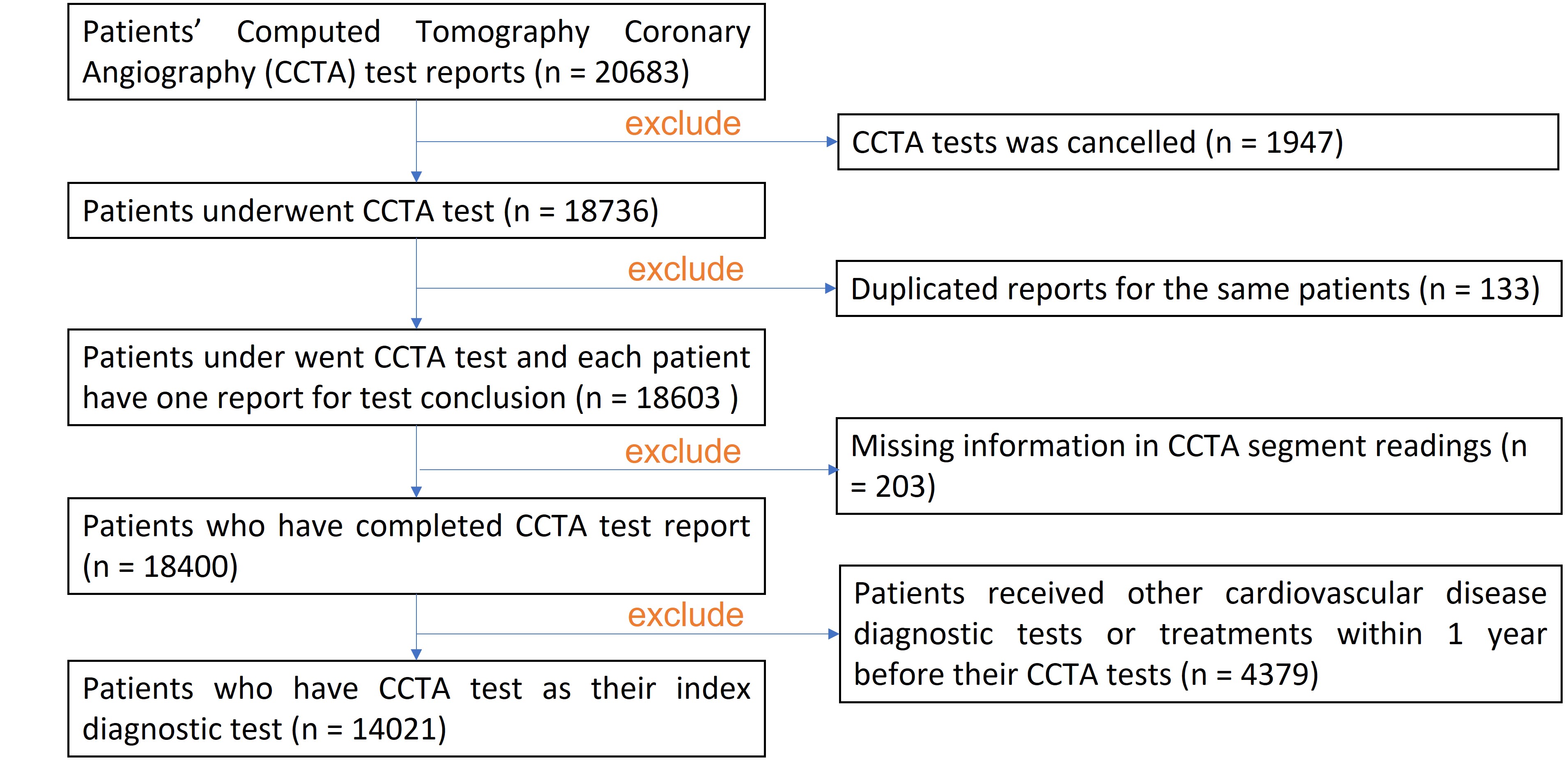}}
\caption{Cohort selection based on patients' CCTA test report availability and investigation journey.}\label{Cohortselection}
\end{figure}

We used a substantial dataset acquired from the University of Ottawa Heart Institute, Canada. 
In the total cohort, 20,683 CCTA test reports were identified in the period between 2006 to 2017. The date of their first CCTA test was defined as $t_0$. 
The timeline is presented in Fig.~\ref{Timeline}. 
We excluded patients whose tests were cancelled or whose CCTA readings were missing. Additionally, we excluded duplicate reports for the same patients and kept records from experienced senior readers. 
Our cohort was further restricted to include patients who had CCTA as their primary test, and had not undergone revascularization or other tests (ICA, SPECT, PET, EST) within 1 year before $t_0$. 
More specifically, we included patients with i) CCTA as their initial test, ii) underwent at least one of the SPECT, PET, EST, ICA or revascularizations (percutaneous coronary intervention (PCI) and coronary artery bypass graft surgery (CABG)) within one year after $t_0$, and iii) had a second test interpretation available. 
This resulted in the selection of a cohort of 884 patients as showed in Fig.~\ref{Cohortselection}. All reported results were obtained from a held-out test set that was not used to train or tune the model.

\subsection{Predictors}
Each patient entry included demographic information (age, sex, weight, and height), clinical information (heart rate, blood pressure, and reasons for test), pretest probability of CAD (Diamond \& Forrester)~\cite{diamond_analysis_1979}, significant cardiovascular risk factors (family history, smoking status, diabetes mellitus, hypertension, hyperlipidemia, and history of CAD), medications (aspirin, beta-blockers, calcium channel blockers, statins, and metformin) and CCTA segment stenosis readings. 
All of this information was included as input to our models.

\subsection{Pre-processing}
 We removed features that had more than 50\% missing values and imputed the remaining missing values in 22 features with maximum missing rate of 8.8\% using multivariate imputation~\cite{buuren_mice_2011}. All the other input features were used to estimate the missing values in a round-robin fashion. 

\subsection{Outcomes}\label{subsec3}
We linked the diagnostic tests and revascularization data for each individual patient using the available data. The downstream tests or treatments were those performed within one year after their CCTA tests. We separated the down streaming test or treatments into two groups: functional testing (including EST, SPECT and PET) vs. invasive tests and/or revascularization (including ICA, PCI and CABG). This makes the downstream tests or treatments prediction to be a binary classification task.  

\begin{figure}[h]%
\centerline{\includegraphics[width=\columnwidth]{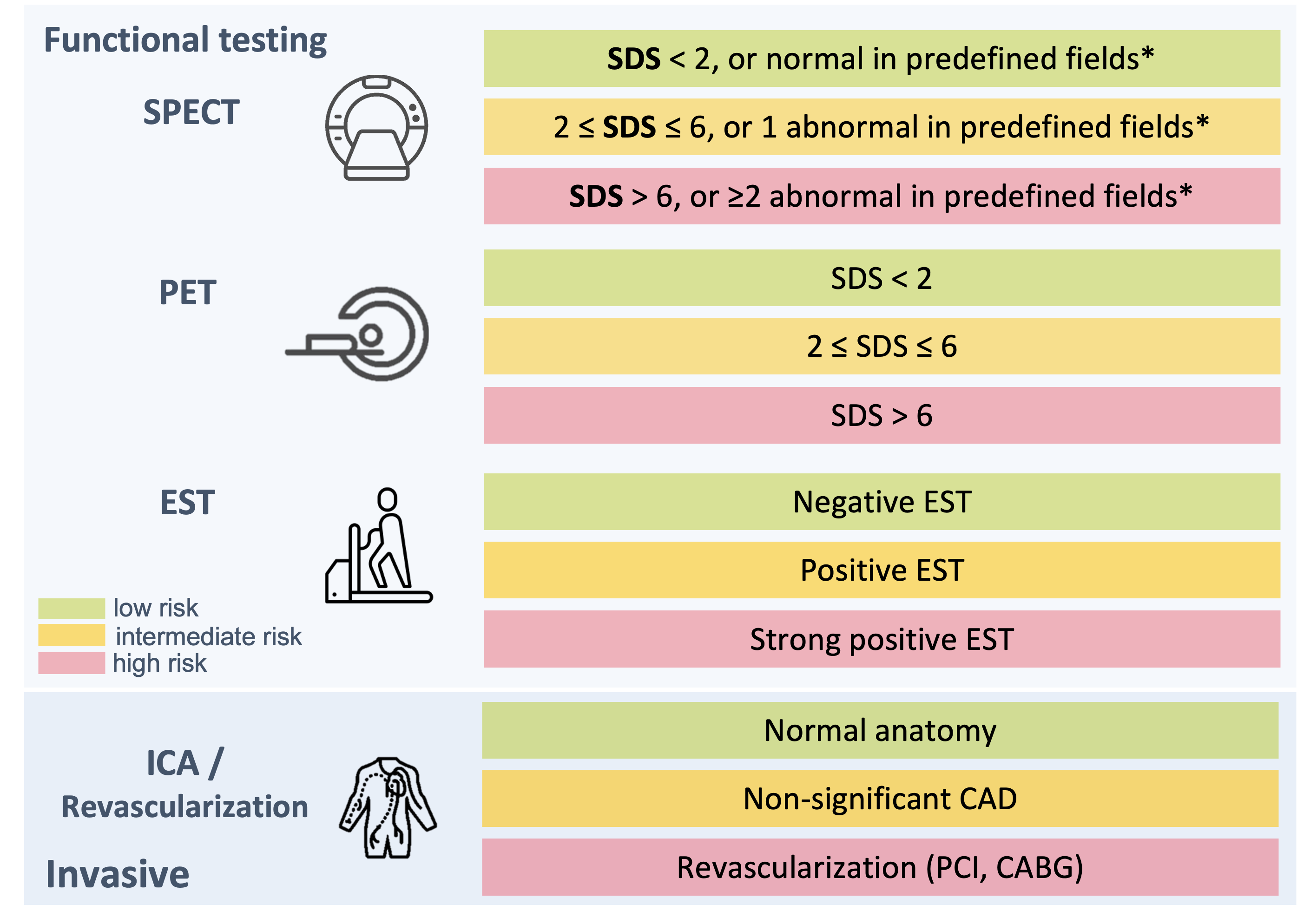}}
\caption{Output labeling. The test conclusions were utilised to determine the risk of CAD for patients who had EST following their CCTA.The Summed Difference Score (SDS) was used to determine the risk of CAD for patients who received SPECT and PET.\cite{czaja_interpreting_2017} This was determined by test findings for patients who had ICA. Patients who were assigned for revascularisation were determined to be high risks. Predefined fields included: scar, ischemia, LAD, RCA, LCX, RCALCX, TID, DilatedLV. \\
LAD, left anterior descending artery; RCA, right coronary artery; LCX, left circumflex artery; TID, transient ischemic dilation; LV, left ventricular; ICA, invasive coronary angiography; PCI, percutaneous coronary intervention; CABG, coronary artery bypass graft surgery}\label{Outputlabel}
\end{figure}

The risk of CAD for patients was determined by the readings of the follow-up tests as shown in Fig.~\ref{Outputlabel}. The risk of patients was classified into three groups: low, intermediate and high risk. Then the risk prediction task becomes a classification task with multiple classes. Based on the tests and data availability, the risks of CAD were defined differently for each test. 

The results of the test were used to determine the risk for patients who had EST after their CCTA. The Summed Difference Score (SDS) was used to determine the risk of CAD for patients who received SPECT and PET~\cite{czaja_interpreting_2017}. This was determined by test findings for patients who had ICA. Patients who were scheduled for revascularisation were determined to be at high risk. 

\subsection{Statistical analysis and metrics for performance measure}
The discriminant power of the risk prediction and downstream test or treatment recommendation tasks was evaluated using the area under the receiver operating characteristic curve (AU-ROC) with a 95\% confidence interval (CI). 
We also reported sensitivity (recall), specificity, precision (positive predictive value) and negative predictive value. The calibration was evaluated using the Brier score, which calculates the difference between the estimated and observed results.
We evaluated the performance of GBDT, multi-layer perceptron (MLP) and TabPerciever in performing single task learning, and also assessed the performance of the multi-task TabPerceiver. 
We then compared the performance of different models using a paired sample T-test and reported the corresponding p-value. The group difference was compared using the chi-square test for categorical variables. One-way ANOVA was used to compare the means of continuous variables across distinct risk groups, whereas the T-test was used to compare down-streaming diagnostic tests or treatments. We regarded the performance difference to be statistically significant if there was no overlap between 95\% CIs and the p-value was less than 0.05. 

\subsection{The architecture of TabPerceiver}
\begin{figure*}%

\includegraphics[width=\textwidth]{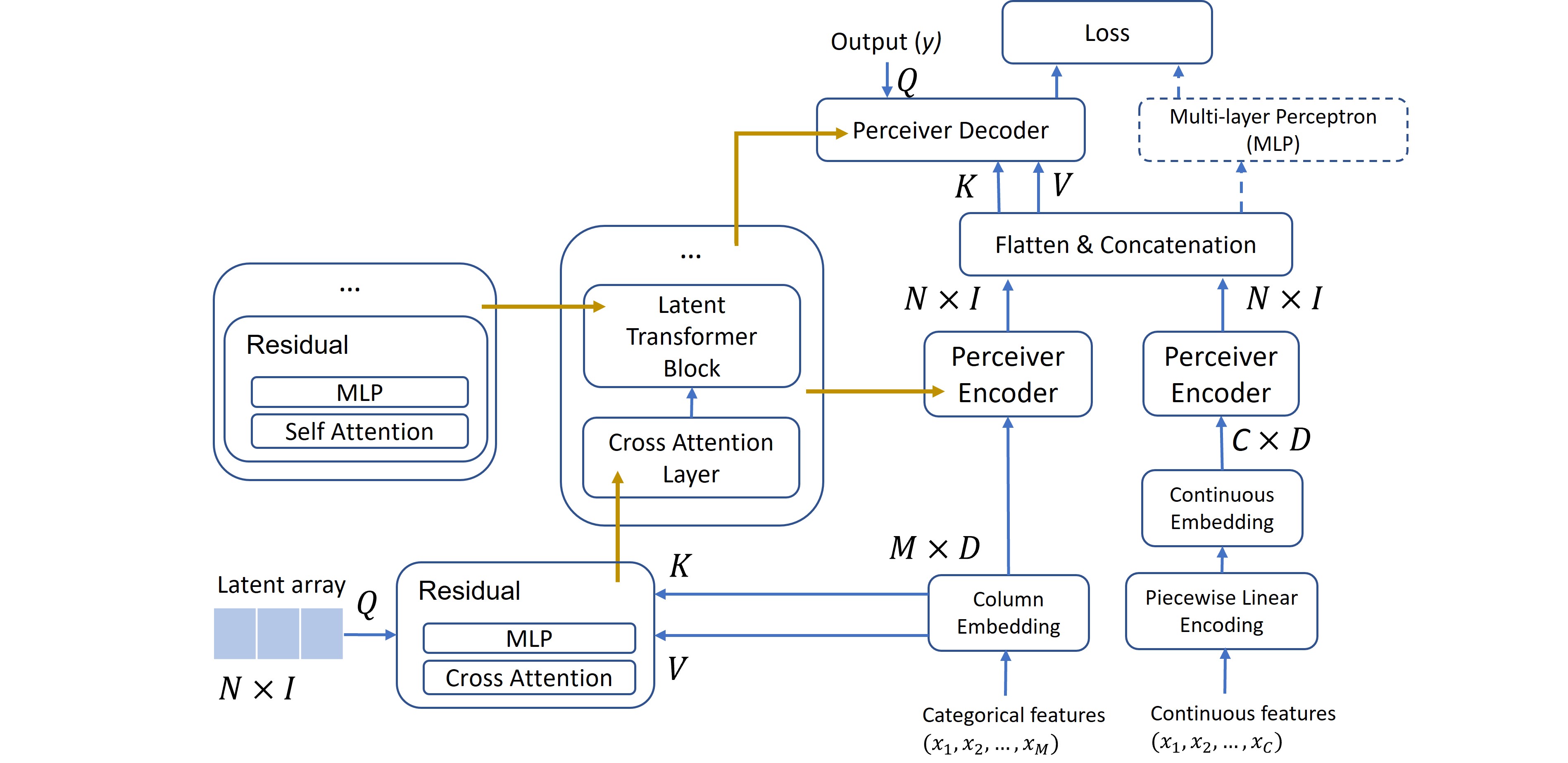}
\caption{The TabPerceiver architecture. It uses a cross-attention module to project the input embeddings to a fixed-dimensional latent bottleneck and then processing it using a stack of Transformer-style self-attention blocks in the latent space. The latent indices $N$ and latent dimension $I$ are hyper-parameters and independent from the number of input features $M$ and $C$. $D$ is the dimension of embedding. The encoder iteratively attends to the input by alternating cross-attention and latent self-attention blocks. The decoder takes the concatenated output of encoder as input to its key ($K$) and value ($V$) networks, and the model output was the input to its query ($Q$) network.}\label{architecture}
\end{figure*}

We have extended the architecture of Perceiver-IO~\cite{jaegle_perceiver_2022} and Tab-transformer~\cite{huang_tabtransformer_2020} to create the Tab-Perceiver. Perceiver-IO is an advanced artificial intelligence architecture that can handle arbitrary inputs and outputs. It can produce various outputs, such as language and video, and is built on the same building blocks as the original Perceiver~\cite{jaegle_perceiver_2021}. It is computationally efficient and can handle large inputs and outputs than the standard self-attention Transformer~\cite{vaswani_attention_2017}.
We constructed separate embeddings for categorical and continuous features implementing different algorithms. The categorical and continuous embeddings were then fed into perceiver encoders individually. Perceiver encoder is a stack of cross-attention modules that transfer the input embedding to a latent vector, with a latent Transformer~\cite{radford_language_2019} converting one latent vector to another latent vector. The latent transformer consists of a multi-head self-attention layer. The perceiver decoder is constructed in a similar manner to the encoder. Both encoder and decoder take in two inputs. The first input is used as the input to the key and value network of the cross-attention module, while the second is used as the input to the query network. The architecture of TabPerceiver is shown in Fig.~\ref{architecture}. 

Let $(x, y)$ denote a feature-target pair, where $x \equiv \{x_{cat}, x_{num}\}$. 

$x_{num} \in \mathbb{R}^C$ denotes all of the $C$ continuous features. 
Let $x_{cat} \equiv \{x_1, x_2, \cdots, x_M\}$ with each $x_i$ being a categorical feature, for $i \in \{1, \cdots, M\}$.
\noindent
We embedded each of the $x_i$ categorical features into a parametric embedding of dimension $D$ using feed-forward network with a learnable embedding layer.  Let $e_{\phi_i} (x_i) \in \mathbb{R}^c$ for $i \in \{ 1, \cdots, M\}$ be the embedding of the $x_i$ feature, and $E_\phi(x_{cat}) = \{e_{\phi_1}(x_1), \cdots, e_{\phi_M}(x_M)\}$ be the set of embeddings for all the categorical features. 

Next, these parametric embeddings $E_\phi(x_{cat})$ are input to the perceiver encoder. The perceiver encoder alternatively employed the cross-attention and the Transformer (self-attention) modules. Both cross-attention and self-attention modules are organised around the use of query-key-value (QKV) attention. QKV attention applies three networks - the query, key, and value networks, which are typically MLPs. To each element of $E_{\phi_i}(x_{cat})$, producing three vectors that preserve the index dimensionality $M$. 

To understand how this works in self-attention, it is import to first note that for $Q \in \mathbb{R}^M{}^\times{}^D$, $K \in \mathbb{R}^M{}^\times{}^D$, and $V \in \mathbb{R}^M{}^\times{}^D$, (where $D$ is the dimension of embedding) the complexity of the QKV attention operation - essentially, $softmax(QK^T)V$ - is $\mathcal{O}(M^2)$, as it involves two matrix multiplications with matrices of dimension $M$. In cross-attention, while $K$ and $V$ are projections of the input, $Q$ is a projection of a learned latent array with index dimension $N$, where the latent's dimension $N$ is a hyper-parameter. Consequently, the cross-attention operation has complexity $\mathcal{O}(MN)$.

The output of the cross-attention module takes the shape of the $e_{\phi_i}(x_i)$ to the $Q$ network. Hence, the cross-attention layer introduces a bottleneck. We can then further analyze this bottleneck in depth by utilizing multi-head self-attention module in the latent space, and the this comes at the cost of $\mathcal{O}(N^2)$. This results in an architecture with complexity $\mathcal{O}(MN + LN^2)$, where $L$ represents the number of layers of self-attention module in perceiver encoder. The introduction of $N$ decouples the input size from the depth of self-attention module, allowing for more flexibility in terms of the input size and depth of the model. For large-scale data with high dimension $M$, $N << M$, helps to reduce the complexity of the model. This allows for construction of deep networks with a lower cost. $N > M$ is also acceptable for small-scale tabular data with selected input features. It contributes to the expansion of input for subsequent layers.

The bottleneck may impede the network's ability to acquire detailed information from the input. To mitigate this impact, the perceiver encoder could be structured with multiple cross-attend layers that enable the latent vector to extract information from the input in an iterative manner, which helps to gather more detailed information from the input. Given that there is no ordering of the categorical features in tabular data, positional encoding is not applied.
We denote the sequence of cross-attend layers as a function $f_\theta$. The function $f_\theta$ operates on parametric categorical embeddings $\{e_{\phi_1}(x_1), \cdots, e_{\phi_M}(x_M)\}$ and returns the corresponding latent vector with shape $N\times I$, where the latent's channel $I$ is another hyper-parameter. 

Concurrently, continuous features were first encoded using piece-wise linear encoding~\cite{gorishniy_embeddings_2022}. Let $x_{num} \equiv \{x_1, x_2, \cdots, x_C\}$ with each $x_j$ being a continuous feature, for $j \in \{1, \cdots, C\}$. For each $x_j$, we split its value range into the disjoint set of $T$ bins according to their empirical quantiles, denoted as $B_1, \cdots, B_T$. 
\begin{equation}
B_t = [b_{t-1}, b_t)\,,
\label{eq_bin}
\end{equation}
where $b_t$ is calculated using the empirical quantile function. Trivial bins of zero size are removed. Then we allocated one trainable embedding $e_t$ for each bin $B_t$ and aggregated the embeddings of its bins with $v_t$ as weights, plus bias $v_0$. The embedding for each continuous feature $x_j$ can thus be denoted as:
\begin{equation}
e_t = 
\begin{cases}
0, & x < b_{t_1}, t > 1\\
1, & x \geq b_t, t < T\\
\frac{x-b_t-1}{b_t-b_{t-1}}, & \text{otherwise}
\end{cases}\,,
\end{equation}
\begin{equation}
e_{\varphi_j}(x_j) = v_0 + \sum_{t=1}^{T}e_t\cdot v_t  \,,
\label{eq_num_emb}
\end{equation}
and where $E_\varphi(x_{cont}) = \{e_{\varphi_1}(x_1), \cdots, e_{\varphi_C}(x_C)\}$ is the set of embeddings for all the continuous features. 
\noindent
The latent vector of categorical features are then flattened and concatenated with the latent vector of continuous features to form a vector of dimension $2NI$. This vector is one of the input to perceiver decoder. We denote the decoder as a function $g$, and it also operates on the target $y$. Let $H$ be the cross-entropy for classification tasks. We minimize the following loss function $\mathcal{L}(x,y)$ to learn all the TabPerceiver parameters. The TabPerceiver parameters include $\phi$ for column embedding, $\theta$ for cross-attend layers. \\
\begin{equation}
\mathcal{L}(x,y) \equiv H(g(f_\theta(E_\phi(x_{cat}), f_\theta(E_\varphi(x_{num})), y))\,. 
\label{tabperceiver}
\end{equation}

As an alternative to decoder, we investigated the use of MLP in the architecture. After undergoing embedding and perceiver encoder, the concatenated latent output of categorical and continuous embeddings can simply be passed through an MLP denoted as $k_\psi$ to predict the target $y$. The equation~\eqref{tabperceiver} can be updated as below, if the MLP with $\psi$ for the top layer were used after the encoder. 
\begin{equation}
\mathcal{L}(x,y) \equiv H(k_\psi(f_\theta(E_\phi(x_{cat})), f_\theta(E_\varphi(x_{num}))), y)\,.
\label{tab_enc_mlp}
\end{equation}

Hard and soft parameter sharing are two prevalent techniques for performing multitask learning in deep neural networks. 
Hard parameter sharing maintains task-specific output layers while sharing the hidden layers between all tasks. Soft parameter sharing preserves individual hidden layers for each task. ~\cite{ruder_overview_2017}
In our multitask TabPerceiver, we opted for the hard parameter sharing strategy, where we shared the hidden layers between all tasks, but kept task-specific output layers. We had two tasks during the learning, risk and downstream tests or treatments prediction, denoted $y^t \in \mathbb{R}^N, t \in \{1,2\}$. We employed one decoder for each task, and then we applied uncertainty to weight the loss~\cite{kendall_multi-task_2018}, where the weight $w_i$ for each task is a learnable parameter. Then the equation for multi-task TabPerceiver would form as: 
\begin{equation}
\mathcal{L}(x,y) \equiv \sum\limits_{i=1}^{T}w_i\mathcal{L}(x, y^i)\,.
\label{eq6}
\end{equation}
\noindent

\subsection{Feature selection}
Feature ranking in GBDT was obtained by computing Shapley Additive Explanation values (SHAP)~\cite{lundberg_unified_2017}. We then selected the top ranked features and fitted them in the GBDT. We conducted an ablation study to assess the importance of different features in TabPerceiver (demographics, clinical information, comorbidity history, medications, CCTA scan readings) by alternatively removing each of these features. We also experimented with the selected features in GBDT on the TabPerceiver model. The detailed feature and their selection results can be found in Appendix section "List of available and selected features".  

\subsection{Hyper-parameter tuning}
Bayesian optimization~\cite{shahriari_taking_2016} was used to tune the model hyper-parameters for all the experimental models, including learning rates, batch size, bins count for continuous feature encoding, number of heads, cross-attend layers and latent dimension in TabPerceiver. We also tuned the number of estimators, number of leaves, maximum depth, minimum data in each leaf, maximum bin and subsample in GBDT. All the models were tuned on the validation set, and the optimal hyper-parameter setting for ML-TabPerceiver is listed in the Appendix section "Model hyper-parameters".

\section{Results}
\subsection{Patient characteristic}
\begin{figure}[h]%

\includegraphics[width=\columnwidth]{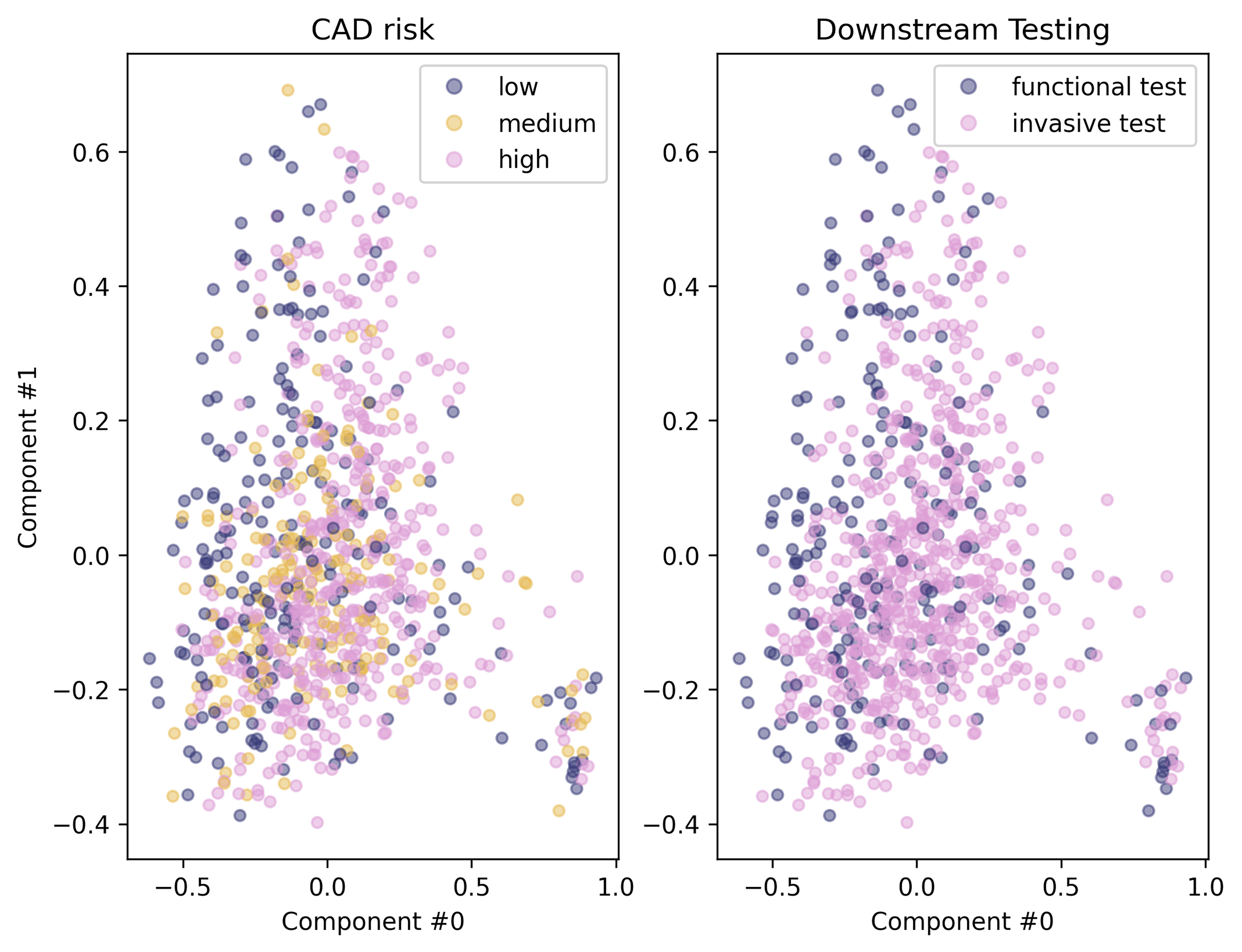}
\caption{Task profile. CAD risk and down-streaming tests are visualized separately using kernalized Principal Component Analysis with 2 principal components. Classes are separated by colors. Each point is a patient entry. }\label{task_corr}
\end{figure}

Out of the 894 patients in the cohort, 475 (53.7\%) were determined to be high risk, 225 (25.5\%) were found to be low risk, and around one-fifth of the patients, 184 (20.8\%), were defined as intermediate risk. Of the total number of patients, 239 (27\%) received functional testing as a downstream diagnostic test, while the remaining 645 (73\%) received invasive testing or revascularization. The mean age was 59.0 years (SD 10.8) and 65.6\% were male (Table~\ref{tab1}). Both hyperlipidemia (61.8\%) and hypertension (56.3\%) were prominent risk factors, whilst diabetes mellitus was presented in 16.2\% of the population and 16.9\% were current smokers. The mean baseline heart rate was 67 (SD 11.9), mean Agatston score was 361.3 (SD 551.4), and the mean baseline systolic and diastolic blood pressure were 137 (SD 19.7) and 79 (SD 10.4), respectively. As showed in Fig.~\ref{task_corr}, the CAD risks of patients are highly correlated with their down-streaming tests. Most of patients at high risks of CAD had invasive tests or treatments after their CCTA.

\begin{table}[t]

\caption{Patient characteristics}\label{tab1}%
\setlength{\tabcolsep}{3pt}
\begin{tabular}{@{}p{3cm}p{2cm}p{1.5cm}p{1.5cm}@{}}
\hline
Patient characteristics & n(\%) unless otherwise specified & p-value (CAD risk) & p-value (down-stream tests)\\
\hline
Age, mean (SD), years    &  59.0 (10.8)  & $<$ 0.05  & $<$ 0.05\\
Male   & 580 (65.6)  & 0.49 & 0.32\\
Height (cm), mean (SD)    & 170.0 (10.5)   & 0.98  & 0.14\\
Weight (kg), mean (SD)    & 85.3 (18.4)   & 0.66  & 0.45   \\
Baseline HR (bpm), mean (SD) & 67 (11.9) & $<$ 0.05 & 0.08 \\
Baseline Systolic BP (mmHg), mean (SD) & 137.0 (19.7) & 0.07 & 0.41 \\ 
Baseline Diastolic BP (mmHg), mean (SD) & 79 (10.4) & 0.34 & 0.52\\ 
Agatston, mean (SD) & 361.3 (551.4) & $<$ 0.05 & $<$ 0.05\\
Probability of CAD, mean (SD) & 35.6 (31.2) & $<$ 0.05 & $<$ 0.05\\
Family history of CAD  & 450 (50.9)   & 0.38  & 0.29 \\ 
Hypertension    & 498 (56.3)   & $<$ 0.05 &  $<$ 0.05\\
Hyperlipidemia    & 546 (61.8)   & 0.37 &  0.10\\
Diabetes mellitus    & 143 (16.2)   & $<$ 0.05 & 0.07  \\
Smoking    & 149 (16.9)   & 0.45 &  $<$ 0.05 \\
Test for chest pain reason    & 501 (56.7)   & $<$ 0.05 & $<$ 0.05 \\
Test for equivocal reason  & 66 (7.5)   & 0.09 & 0.81  \\
Dyspnea    & 547 (61.9)   & 0.15 & $<$ 0.05\\
Palpitations    & 414 (46.8)   & $<$ 0.05 & 0.44 \\
Syncope    & 280 (31.7)   & 0.18 & 0.83 \\
Aspirin   & 514 (58.1)   & $<$ 0.05 & $<$ 0.05 \\
ACE inhibitors  & 240 (27.1)   & $<$ 0.05 &  $<$ 0.05 \\
Beta-blockers  & 390 (44.1)   & $<$ 0.05 & $<$ 0.05 \\
Calcium Channel Blockers    & 132 (14.9) & 0.24 & $<$ 0.05 \\
Diuretics    & 157 (17.8)   & 0.57 & 0.06  \\
Statins    & 458 (51.8)   & 0.10 & 0.07 \\
\hline
\multicolumn{4}{p{249pt}}{CAD, coronary artery disease; SD, standard deviation; HR, heart rate; BP, blood pressure; ACE, angiotensin-converting enzyme}\\
\end{tabular}
\end{table}

\subsection{Model performance using the selected features}
Table~\ref{tab2} shows the AUC, sensitivity, specificity and brier score of the multi-task TabPerceiver with decoder using the selected input variables comparing the performance of single-task GBDT, MLP and TabPerceiver. GBDT outperformed the competitors while doing the single task learning. However, multi-task TabPerceiver showed the best discrimination and calibration capabilities among the evaluated models in the CAD risk prediction task. It also achieved optimal performance in predicting the downstream diagnostic tests.  

Regarding the selection of MLP and decoder within the TabPerceiver architecture, the decoder performed marginally better when used as the output layer.

\begin{figure}[b]%

\includegraphics[width=\columnwidth]{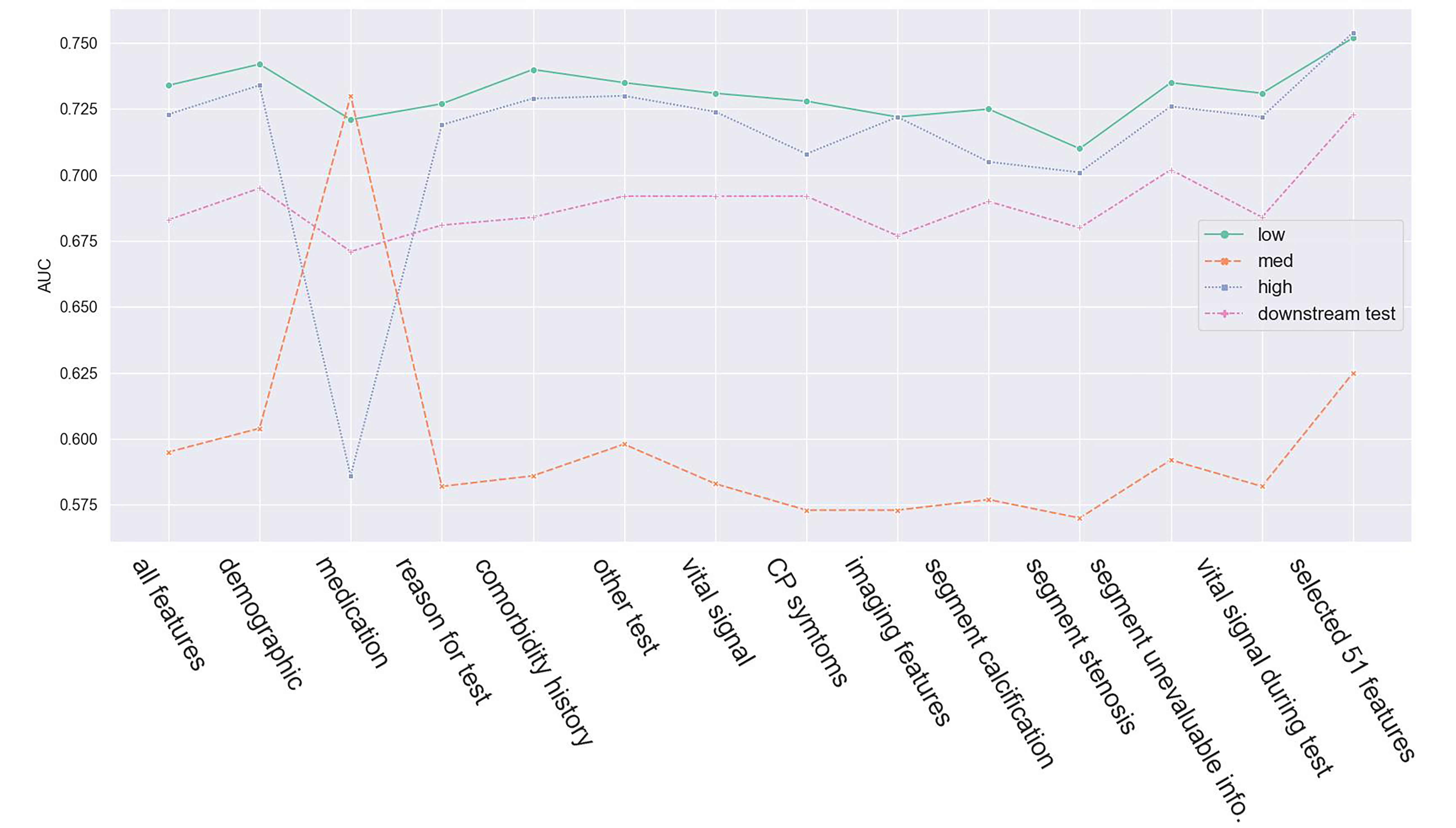}
\caption{Ablation study with exclusion of different feature modalities by CAD risk group and downstream tests. CAD, coronary artery disease, CP, chest pain, info., information.}\label{ablation}
\end{figure}

\begin{table*}

\caption{Performance comparison among different machine learning methods}\label{tab2}
\begin{tabular*}{\textwidth}{@{\extracolsep{\fill}}lccccc@{\extracolsep{\fill}}}
\hline%
\textbf {Models} & \textbf{Target} & \multicolumn{4}{@{}c@{}}{\textbf{Performance}}  \\
&& AUC (95\% CI) & Sensitivity & Specificity & Brier Score\\
\hline%
\multirow{3}{*}{GBDT}&low & 0.733 (0.721, 0.747) & 0.58 & 0.83 & 0.31\\
&medium& 0.602 (0.589, 0.616) & 0.18 & 0.81 & 0.25\\
&high& 0.719 (0.706, 0.732) & 0.68  & 0.63 & 0.34\\
&invasive test& 0.730 (0.715, 0.745) & 0.64 & 0.42 & 0.34\\
\hline
\multirow{3}{*}{MLP}&low & 0.685 (0.665, 0.705) & 0.51 & 0.82 & 0.32\\
&medium& 0.593 (0.577, 0.609) & 0.19 & 0.80 & 0.29\\
&high& 0.696 (0.676, 0.716) & 0.65  & 0.57 & 0.35\\
&invasive test& 0.694 (0.681, 0.706) & 0.70 & 0.71 & 0.33\\
\hline
\multirow{3}{*}{TabPerceiver (MLP)}&low & 0.723 (0.711, 0.734) & 0.54 & 0.83 & 0.31\\
&medium& 0.602 (0.590, 0.615) & 0.19 & 0.80 & 0.27\\
&high& 0.713 (0.702, 0.725) & 0.68 & 0.63 & 0.34\\
&invasive test & 0.704 (0.693, 0.681) & 0.69 & 0.41 & 0.33\\
\hline
\multirow{3}{*}{TabPerceiver (decoder)}& low &0.717 (0.705, 0.729) & 0.53 & 0.83 & 0.30\\
&medium& 0.588 (0.574, 0.601) & 0.17 & 0.80 & 0.27\\
&high& 0.703 (0.690, 0.715) & 0.68 & 0.62 & 0.36\\
&invasive test & 0.688 (0.677, 0.699) & 0.72 & 0.42 & 0.33\\
\hline
\multirow{3}{*}{ML-TabPerceiver (MLP)}&low & 0.723 (0.711, 0.734) & 0.54 & 0.83 & 0.31\\
&medium& 0.602 (0.590, 0.615) & 0.19 & 0.80 & 0.27\\
&high& 0.713 (0.702, 0.725) & 0.68 & 0.63 & 0.34\\
&invasive test & 0.704 (0.693, 0.681) & 0.69 & 0.41 & 0.33\\
\hline

\multirow{3}{*}{ML-TabPerceiver (decoder)}&low & \textbf{0.752} (0.742, 0.763) & 0.59 & 0.85 & 0.29\\
&medium& \textbf{0.625} (0.611, 0.640) & 0.22 & 0.81 & 0.27\\
&high& \textbf{0.754} (0.744, 0.764) & 0.70 & 0.66 & 0.31\\
&invasive test & 0.723 (0.711, 0.735) & 0.71 & 0.44 & 0.32\\
\hline

\end{tabular*}
GBDT, gradient boosting decision tree; MLP, multi-layer perceptron; ML, multi-task learning; AUC, area under the receiver operating characteristic curve; CI, confidence interval
\end{table*}

\subsection{Feature Importance}
\begin{figure*}%
\centering
\includegraphics[width=0.9\textwidth]{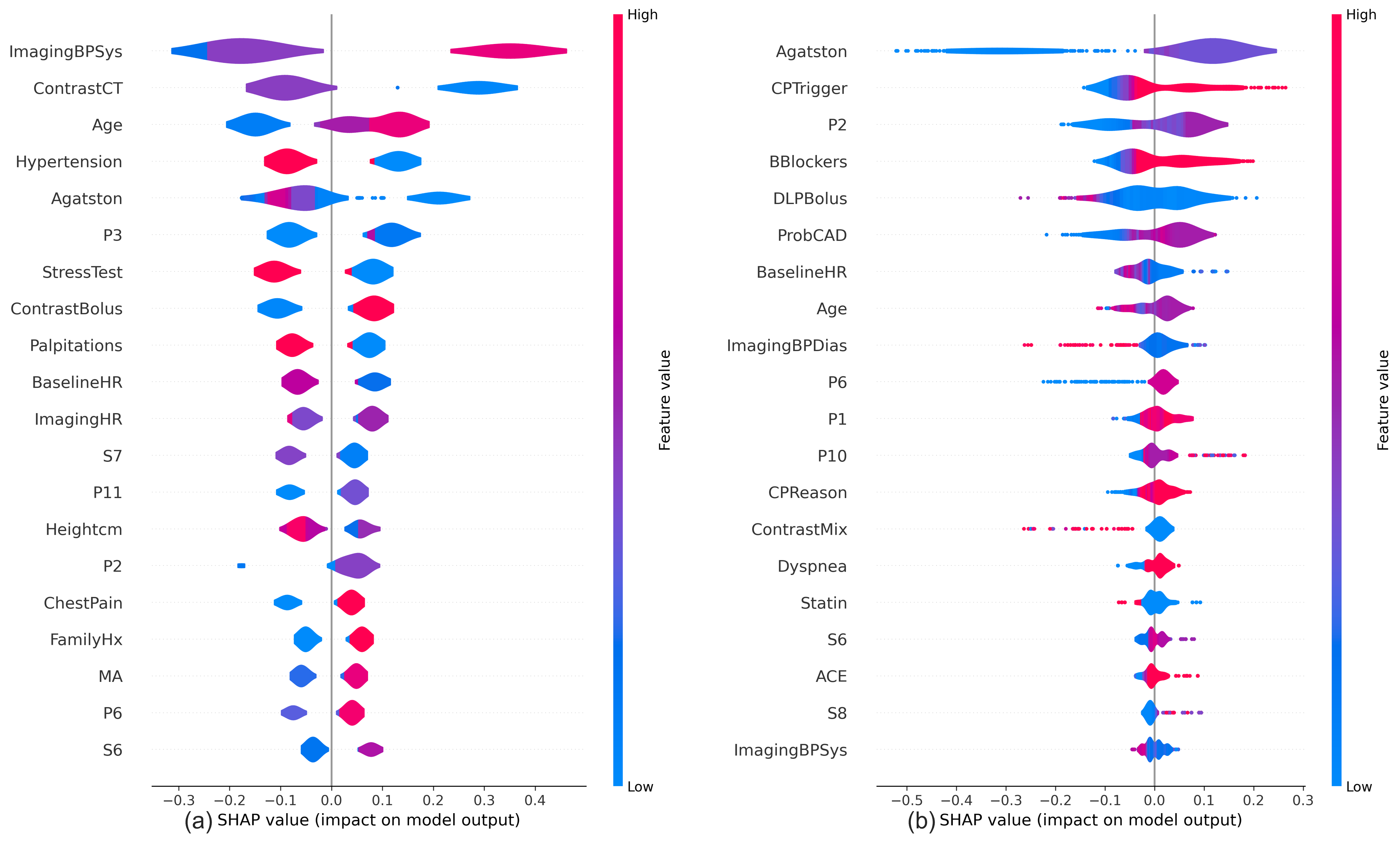}
\caption{Feature importance for predicting risk of CAD (a) and down-streaming diagnostic test (b). Blood pressure, age,  hypertension and segment readings are ranked as the top important features for predicting patients' risk of CAD. Agatston score is the most determinative feature for down-streaming test selection. Other important features include blood pressure, segment readings, beta-blocker and chest pain trigger. CT, computed tomography, HR, heart rate, FamilyHx, family history, P3, calcified level of distal right coronary, S7, stenosis level of mid left anterior descending artery, P11, calcified level of proximal left circumflex artery, P2, calcified level of mid right coronary artery, MA, milliampere-seconds, P6, calcified level of proximal left anterior descending artery, S6, stenosis level of proximal left anterior desending artery, CP, chest pain; DLP, dose length product; P2, calcified level of mid right coronary artery; ProbCAD, pre-test probability of CAD (Diamond \& Forrester); BPDias, diastolic blood pressure; P1, calcified level of proximal right coronary artery; P10, calcified level of the second diagonal artery;  ACE, angiotensin converting enzyme inhibitors; S8, stenosis level of distal left anterior descending artery; BPSys, systolic blood pressure}\label{feature_imp}
\end{figure*}

Fig.~\ref{ablation} shows the results of the ablation study. It demonstrates that segment readings such as stenosis and calcification degrees have a stronger impact on model performance than other feature modalities, such as demographical characteristics and other tests received by the patient. Multi-tasking Tabperceiver achieved the optimized performance on the selected features from different feature modalities. Feature ranking obtained using SHAP turned out to be the most effective guide for feature selection in both GBDT and deep learning models for CAD risk and down-streaming test prediction. Fig.~\ref{feature_imp} shows the feature importance ranking by their SHAP values in the GBDT model. Clinical information such as blood pressure and heart rate are revealed to be important predictors for both tasks, as well as the Agatston score which relate to the result and the confidence in the interpretation of the CCTA test. Similar to the previous studies, age and hypertension were found to be significant risk factors for CAD~\cite{alaref_machine_2020, nowbar_mortality_2019}. CCTA segment readings and pre-test probability also had an important role in predicting both tasks. This was also consistent with the results of the ablation study.

\section{Discussion}

Coronary artery disease is a commonly encountered disease associated with significant morbidity, mortality, and high medical costs. It has been common in clinical practice to deploy a validated pretest probability model for CAD to guide the diagnostic test selection. The majority of existing tools primarily focus on the selection of gatekeeper tests, whereas only a handful of systems contain information on downstream testing or treatment~\cite{morgan-hughes_downstream_2021, fyyaz_computed_2020}.
Hence, there is a need for models that are based on clinical observations, which can predict the need for downstream tests or treatments in order to identify patients who are at high risk and would benefit from invasive diagnostic tests or revascularization. This study used patient clinical characteristics and CCTA segment readings to predict the risk of CAD and selection of downstream tests or treatments.
We employed a transformer-based tabular deep learning model as a novel optimized analytical approach to create clinical decision support models. Our results show that the developed multi-task TabPerceiver model is effective in predicting the risk of CAD and the selection of downstream tests or treatments. Additionally, we found that the model has appropriate calibration (the predicted probabilities of outcome reflect true probabilities of the outcome).

The use of the ML model in a clinical setting could streamline the identification of patients who may benefit from invasive diagnostic evaluation and revascularization, reducing the need for time-consuming routine clinical steps.
Meanwhile, the overuse of diagnostic imaging modalities is a major issue causing significant stress on the health budget~\cite{muskens_overuse_2022, huang_overuse_2015}. 
To address this, there has been an increased focus on the application of risk stratification and pre-test probability assessment prior to initiation of downstream testing~\cite{fihn_2014_2014, task_force_members_2013_2013}.
Despite this, multiple studies have demonstrated that these risk assessment models performed sub-optimally in certain cohorts~\cite{kumamaru_overestimation_2014, baskaran_comparison_2019}. In recent years, various models have been validated in multiple external populations, with a trend towards a decline in discriminative capacity~\cite{he_diagnostic_2017, alaref_machine_2020}. 
Indeed, differences in derivation (utilization of various imaging modalities as well as different cut-off values for the definition of CAD risk, adoption of various data processing methodologies), model complexity and inconsistent external validation limit their usefulness in clinical practice. In an ever-changing environment where populations are longitudinally evolving due to changes in dietary habits, environmental exposures, advances in science and technology, it is essential to have models that adapt over time. To address this need, machine learning has been increasingly used in the cardiovascular field. 
Machine learning involves algorithms that are specifically designed to identify relationships between data that go beyond the traditional linear statistical approaches. 

Moreover, machine learning leverages the growing accessibility of computational power and storage space to rapidly produce meaningful results based on complex input data. Furthermore, machine learning has proven itself to be a powerful tool for diagnosing coronary artery disease~\cite{alizadehsani_machine_2019}. We further demonstrate the robustness of TabPerceiver on noisy data, in comparison to the baseline MLP. We evaluated these two scenarios on categorical and continuous features to specifically demonstrate the robustness of contextual embeddings from the Perceiver encoder layers.

In our multitask learning framework, the Perceiver encoder was devised to represent task-specific features. The model with common-task neurons, which comprised approximately 20\% of the perceiver layer, achieved the best performance. The introduction of common-task layers into the perceiver layer can assist in learning the shared complementary information for different tasks. These findings demonstrate that multitask learning can enhance the prediction performance of the correlated tasks. Additionally, it has the advantage of improving data efficiency and reducing over-fitting through shared representations. This can be attributed to the fact that prediction with the associated task can reduce model overfitting and the impact of noisy information from the heterogeneous tumor by learning related and complementary features.

The deep feature/prediction from multiple tasks in the TabPerceiver was fused using three information fusion strategies. The results indicate the effectiveness of the methods compared to the GBDT. The experimental results show the superior performance of our multitask-learning-based model over single-task learning. 

Feature concatenation may enhance
the combination of complementary features while also replicating information from various sources of categorical and continuous features.

The limited sample size and input seem to hinder the predictive performance of all the models for patients with moderate(medium) risk of CAD. This is consistent with prior findings that current information remains limited for clinicians to accurately determine risk for patients in intermediate risk ranges~\cite{pasternak_task_2003}. In the risk prediction task, we chose the class with the highest probability. Hence, the sensitivity of predicting low and moderate risk was low as expected, since the low prevalence of events has affected the sensitivity and specificity. 

It is important to note that validating the proposed model with tuned hyper-parameters on an additional dataset is of vital importance for the generalization and clinical applications of the model. However, medical imaging tabular data from different datasets/cohorts are usually distinct from each other. This may induce bias and inconsistent results for the tabular data-based model. It would be valuable to conduct further research to assess our design strategy when a large dataset becomes available in the future.

In our single task learning experiment, the deep models performed poorly on small tabular datasets and were no better than GBDT. We proposed the use of multi-task TabPerceiver using feature embeddings and an attention-based module. This architecture performed better than the single-task deep learning model we experimented with. We also examined the trade offs between various architecture, computational inference cost, and hyper-parameter optimization time, which are important in real-world applications. Our analysis demonstrates that the deep learning models are more sensitive to input features. As showed in Table.~\ref{fs_impact}, the prediction accuracy of GBDT is not significantly impacted by feature selection, however, removing uninformative features has improved the performance of deep learning models.
This has also reported in the recent study~\cite{grinsztajn_why_2022}.  Additionally, it is much more challenging to optimize deep models than GBDT on a new dataset, due to their large number of hyper-parameters and high computational cost. However, we discovered that TabPerceiver performed effectively on tabular data while incurring a lower computational burden.
 
\begin{table}
\caption{Impact of feature selection on model performance}\label{fs_impact}
\setlength{\tabcolsep}{3pt}
\begin{tabular}{@{}p{2cm}p{1.5cm}p{2cm}p{2.5cm}@{}}
\hline%

\textbf{Models} & \textbf{Number of input features} & \textbf{Target} & \textbf{AUC (95\% CI)}\\
\hline%
\multirow{8}{*}{GBDT} & \multirow{4}{*}{146} & low &0.733 (0.721, 0.747) \\
&&medium& 0.602 (0.589, 0.616) \\
&&high& 0.719 (0.706, 0.732) \\
&&invasive test& 0.730 (0.715, 0.745) \\
\cline{2-4}
&\multirow{4}{*}{51} & low &0.726 (0.712, 0.740) \\
&&medium& 0.610 (0.595, 0.626) \\
&&high& 0.715 (0.701, 0.729) \\
&&invasive test& 0.741 (0.731, 0.752) \\
\hline
\multirow{8}{*}{MLP} & \multirow{4}{*}{146} & low &0.660 (0.649, 0.672) \\
&&medium& 0.580 (0.571, 0.590) \\
&&high& 0.672 (0.661, 0.684) \\
&&invasive test& 0.669 (0.656, 0.682) \\
\cline{2-4}
&\multirow{4}{*}{51} & low &0.685 (0.665, 0.705) \\
&&medium& 0.593 (0.577, 0.609) \\
&&high& 0.696 (0.676, 0.716) \\
&&invasive test& 0.694 (0.681, 0.706) \\
\hline
\multirow{8}{*}{TabPerceiver} & \multirow{4}{*}{146} & low &0.689 (0.671, 0.702) \\
&&medium& 0.573 (0.556, 0.592) \\
&&high& 0.631 (0.605, 0.658) \\
&&invasive test& 0.680 (0.669, 0.692) \\
\cline{2-4}
&\multirow{4}{*}{51} & low & 0.700 (0.687, 0.712) \\
&&medium& 0.593 (0.585, 0.607) \\
&&high& 0.673 (0.663, 0.682) \\
&&invasive test& 0.685 (0.676, 0.704) \\
\hline
\multirow{8}{*}{ML-TabPerceiver} & \multirow{4}{*}{146} & low &0.734 (0.724, 0.745) \\
&&medium& 0.595 (0.579, 0.610) \\
&&high& 0.723 (0.711, 0.735) \\
&&invasive test& 0.683 (0.664, 0.702) \\
\cline{2-4}
&\multirow{4}{*}{51} & low & 0.752 (0.742, 0.763) \\
&&medium& 0.629 (0.616, 0.643) \\
&&high& 0.756 (0.746, 0.767) \\
&&invasive test& 0.730 (0.715, 0.745) \\
\hline

\end{tabular}
GBDT, gradient boosting decision tree; MLP, multi-layer perceptron; ML, multi-task learning; AUC, area under the receiver operating characteristic curve; CI, confidence interval
\end{table}

Follow-up admission and mortality outcome information for the entire cohort is not available in our dataset. We have thus used test readings and interpretation to define the risk of CAD for patients, resulting in a significant reduction in the sample size of the study cohort. This may introduce potential bias to our study and limit the performance of our GBDT and deep learning models, as it has been found to be tailored towards big data. Future validation might be required using a large cohort. The transformer-based model has showed its effectiveness in handling images, audio and text data. The perceiver architecture is also designed to process large input and output. In future work, we would like to incorporate both tabular and imaging data into the model, where further performance enhancement may be expected.

\section{Conclusion and future work}

We propose TabPerceiver, a novel deep learning tabular data modeling architecture, that is easy to use and computationally efficient. Our deep learning model outperforms MLP in single task learning, and the performance of multi-tasking TabPerceiver was comparable to or better than that of the GBDT. It appears that deep learning tabular architectures benefit from multi-task learning. We have also demonstrated that feature selection is an essential step for deep learning tabular data modeling. Together with proper feature embedding (both categorical and continuous feature embedding), these strategies address the drawbacks of deep learning on tabular data. In a future study, we plan to incorporate more data modalities, such as actual CCTA images and clinical text reports, into the model to improve the prediction of risk of CAD and support the selection of downstream diagnostic testing and treatments.
\section{Appendix}
\begin{longtblr}[
caption = {Candidate and selected features},
label={tab:feature_detail},
note{} = {CAD, coronary artery disease; HR, heart rate; CHF, congestive heart failure; BP, blood pressure; ACE, angiotensin-converting enzyme; ASA, acetylsalicylic acid; AT2, angiotensin II type 2 receptors; NOACs, novel oral Anticoagulants; PPI, proton pump inhibitors; CABG, coronary artery bypass graft surgery; Cath, catheterization; PVD, peripheral vascular disease; RP, renal protection; CD, conduct disorder; CT, computed tomography; DLP, dose length product; MA, milliampere-seconds; kV, 
X-Ray tube voltage},
]{
colspec={X}, width=0.5\linewidth,
rowhead = 1,
rowsep = 0pt}
\hline
{Feature\\modality} & Feature & Type & Select\\
\hline
\SetCell[r=6]{m, 1cm} Demo-graphics & Age (year) & cont. & yes\\
& Male  & binary & no\\
& Height (cm) & cont. & no\\
& Weight (kg) & cont. & yes \\
& Smoking & CAT & yes\\
& Probability of CAD & cont. & yes\\
\hline
\SetCell[r=3]{m, 1cm} Vital sign & Baseline HR (bpm) & cont. & yes \\
& Baseline Systolic BP (mmHg) & cont. & no \\
& Baseline Diastolic BP (mmHg) & cont. & no \\
\hline
\SetCell[r=3]{m, 1cm}{Vital sign \\ (imaging) } & Imaging HR (bpm) & cont. & yes \\
& Imaging Systolic BP (mmHg) & cont. & yes\\
& Imaging Diastolic BP (mmHg) & cont. & no \\
\hline
\SetCell[r=6]{m, 1cm} Reason for test & Ventricular tachycardia reason & binary & no \\
& Chest pain reason & binary & yes\\
& Equivocal reason  & binary & no\\
& Rule out CAD reason & binary & yes\\
& Dyspnea reason & binary & no\\
& CHF reason & binary & no\\
\hline
\SetCell[r=5]{m, 1cm} Other tests& Stress test & binary & no\\
& Stress Echo & binary & no\\
& RNA & binary & yes\\
& Myocardial perfusion imaging & binary & yes\\
& Myocardial viability & binary & yes\\
\hline
\SetCell[r=16]{m, 1cm}{Medications}& ACE inhibitors & binary & no\\
& Aspirin & binary & yes\\
& AT2 & binary & no\\
& Beta blockers & binary & yes\\
& Calcium channel blockers & binary & no\\
& Insulin & binary & no\\
& Metformin & binary & no\\
& NOACs & binary & no\\
& Statins & binary & no\\
& Diuretics & binary & no\\
& Ticlopidine & binary & no\\
& PPI & binary & no\\
& Nitrates & binary & no\\
& Other oral hypoglycemics & binary & no\\
& Vasodilators & binary & yes\\
& Warfarin & binary & no\\
\hline
\SetCell[r=20]{m,1cm}{Comorbidity \\ and other history} & Valvular heart disease & binary & no\\
& Valvular repair or replacement & binary & no\\
& Congenital heart disease & binary & no\\
& Dyspnea & binary & no\\
& Hypertension & binary & yes \\
& Congestive heart failure & binary & no\\
& Palpitations & binary & yes\\
& Cardiac implants & binary & no\\
& Renal insufficiency & binary & yes\\
& Hyperlipidemia & binary & no\\
& Prior CABG & binary & yes \\
& Prior Catheterisation & binary & yes \\
& Myocardial infarction & binary & yes\\
& CAD & binary & no\\
& Diabetes mellitus  & binary & no\\
& Family history & binary & yes\\
& PVD & binary & no\\
& Syncope & binary & no\\
& RP & binary & yes\\
& CD & binary & no\\
\hline
\SetCell[r=55]{m,1cm} {Segment\\readings} & Proximal right coronary artery & &\\
& \hspace{3mm} calcified level (P1) & CAT & no\\
& \hspace{3mm} stenosis level (S1) & CAT & yes\\
& Mid right coronary artery & & \\
& \hspace{3mm} calcified level (P2) & CAT & yes\\
& \hspace{3mm} stenosis level (S2) & CAT & no\\
& Distal right coronary artery  & & \\
& \hspace{3mm} calcified level (P3) & CAT & no\\
& \hspace{3mm} stenosis level (S3) & CAT & yes\\
& Posterior interventricular artery & & \\
& \hspace{3mm} calcified level (P4) & CAT & no\\
& \hspace{3mm} stenosis level (S4) & CAT & no\\
& Left main & & \\
& \hspace{3mm} calcified level (P5) & CAT & yes\\
& \hspace{3mm} stenosis level (S5) & CAT & no\\
& {Proximal left anterior descending \\ artery} & & \\
& \hspace{3mm} calcified level (P6) & CAT & yes\\
& \hspace{3mm} stenosis level (S6) & CAT & yes\\
& {Mid left anterior descending \\ artery} & & \\
& \hspace{3mm} calcified level (P7) & CAT & yes\\
& \hspace{3mm} stenosis level (S7) & CAT & yes\\
& {Distal left anterior descending \\ artery} & & \\
& \hspace{3mm} calcified level (P8) & CAT & yes\\
& \hspace{3mm} stenosis level (S8) & CAT & yes\\
& The first diagonal artery (D1)  & & \\
& \hspace{3mm} calcified level (P9) & CAT & yes\\
& \hspace{3mm} stenosis level (S9) & CAT & yes\\
& The second diagonal artery (D2) & & \\
& \hspace{3mm} calcified level (P10) & CAT & no\\
& \hspace{3mm} stenosis level (S10) & CAT & yes\\
& Proximal left circumflex artery & & \\
& \hspace{3mm} calcified level (P11) & CAT & yes\\
& \hspace{3mm} stenosis level (S11) & CAT & yes\\
& The first marginal artery (M1) & & \\
& \hspace{3mm} calcified level (P12) & CAT & no\\
& \hspace{3mm} stenosis level (S12) & CAT & yes\\
& Mid left circumflex artery & & \\
& \hspace{3mm} calcified level (P13) & CAT & no\\
& \hspace{3mm} stenosis level (S13) & CAT & no\\
& The second marginal artery (M2) & & \\
& \hspace{3mm} calcified level (P14) & CAT & yes\\
& \hspace{3mm} stenosis level (S14) & CAT & no\\
& Distal left circumflex artery & & \\
& \hspace{3mm} calcified level (P15) & CAT & no\\
& \hspace{3mm} stenosis level (S15) & CAT & no\\
& Left ventricle & & \\
& \hspace{3mm} calcified level (P16) & CAT & no\\
& \hspace{3mm} stenosis level (S16) & CAT & no\\
& Ramus intermedius & & \\
& \hspace{3mm} calcified level (P17) & CAT & no\\
& \hspace{3mm} stenosis level (S17) & CAT & yes\\
& Other segments & & \\
& \hspace{3mm} calcified level (P18-22) & CAT & no\\
& \hspace{3mm} stenosis level (S18-22) & CAT & no\\
& \hspace{3mm} unevaluable (Uneval1-22) & binary & no\\
\hline
\SetCell[r=3]{m,1cm} Chest pain symptoms & Chest pain trigger & binary & yes\\
& Chest pain relieved & binary & no\\
& Chest pain & binary & yes\\
\hline
\SetCell[r=17]{m,1cm} {Other imaging \\ information} & Dominance & CAT & no\\
& Agatston & cont. & yes\\
& Contrast Bolus & cont. & no\\
& Contrast CT & cont. & yes\\
& Contrast Mix & cont. & no\\
& Contrast Mix Percent & cont. & no\\
& Contrast Mix Rate & cont. & no\\
& Contrast Type & CAT & no\\
& DLP Bolus & cont. & yes\\
& DLP Cardiac & cont. & yes\\
& DLP non contrast & cont. & no\\
& {Left ventricular \\ end-diastolic volume} & cont. & yes\\
& {Left ventricular ejection \\ fraction} & cont. & no\\
& {Left ventricular \\ end-systolic volume} & cont. & no \\
& MA & cont.& yes\\
& kV & cont.& yes\\
& Primary First & binary & yes\\

\hline

\end{longtblr}

\begin{longtblr}[
caption = {Model hyper-parameter},
label={tab:hyperparam},
]{
colspec={X},
rowhead = 1,
rowsep = 0pt}
\hline
Model & Hyper-parameters\\
\hline
\SetCell[r=11]{m, 0.5cm} {ML-TabPerceiver \\ (selected 51 features)} & batch\char`_size=16\\
& num\char`_layers=2 \\
& input\char`_embed\char`_dim=128 \\
& num\char`_cross\char`_attention\char`_heads=4 \\
& num\char`_self\char`_attention\char`_layers\char`_per\char`_block= 4\\
& num\char`_latents=8\\
& num\char`_latent\char`_channels=32\\
& bins\char`_count=150 \\
& weight\char`_decay=0.01\\
& dropout=0\\
& label\char`_smoothing=0.3\\
\hline
\SetCell[r=11]{m, 0.5cm} {ML-TabPerceiver \\ (all features)} & batch\char`_size=32\\
& num\char`_layers=2 \\
& input\char`_embed\char`_dim=128 \\
& num\char`_cross\char`_attention\char`_heads=8 \\
& num\char`_self\char`_attention\char`_layers\char`_per\char`_block= 2\\
& num\char`_latents=16\\
& num\char`_latent\char`_channels=64\\
& bins\char`_count=50 \\
& weight\char`_decay=0.01\\
& dropout=0\\
& label\char`_smoothing=0.3\\
\hline
\end{longtblr}

\bibliographystyle{IEEEtran}
\bibliography{generic-color-brief}
\end{document}